\pdfoutput=1

\documentclass[11pt]{article}

\usepackage[]{EMNLP2022}
\usepackage{multirow}
\usepackage{caption}
\usepackage{subcaption}
\usepackage{amsmath}
\usepackage{amssymb}
\usepackage{times}
\usepackage{latexsym}
\usepackage{comment}
\usepackage{float}
\usepackage{chngpage}

\usepackage[T1]{fontenc}

\usepackage[utf8]{inputenc}

\usepackage{microtype}

\usepackage{inconsolata}

\usepackage{booktabs}
\usepackage{multirow}
\usepackage{graphicx}
\usepackage{xurl}
\title{On the Impact of Temporal Concept Drift on Model Explanations} 

\author{Zhixue Zhao \quad George Chrysostomou \quad Kalina Bontcheva \quad  Nikolaos Aletras \\
        Department of Computer Science, University of Sheffield \\
        United Kingdom \\
\texttt{\{zhixue.zhao, gchrysostomou1, k.bontcheva, n.aletras\}@sheffield.ac.uk}}

\begin{document}
\maketitle
\begin{abstract}
Explanation faithfulness of model predictions in natural language processing is typically evaluated on held-out data from the same temporal distribution as the training data (i.e. synchronous settings). While model performance often deteriorates due to temporal variation (i.e. temporal concept drift), it is currently unknown how explanation faithfulness is impacted when the time span of the target data is different from the data used to train the model (i.e. asynchronous settings). 
For this purpose, we examine the impact of temporal variation on model explanations extracted by eight feature attribution methods and three select-then-predict models across six text classification tasks. Our experiments show that
(i) faithfulness is not consistent under temporal variations across feature attribution methods (e.g. it decreases or increases depending on the method), with an attention-based method demonstrating the most robust faithfulness scores across datasets; and
(ii) select-then-predict models are mostly robust in asynchronous settings with only small degradation in predictive performance.
Finally, feature attribution methods show conflicting behavior when used in FRESH (i.e. a select-and-predict model) and for measuring sufficiency/comprehensiveness (i.e. as post-hoc methods), suggesting that we need more robust metrics to evaluate post-hoc explanation faithfulness.\footnote{Code for replicating the experiments in this study: \url{https://github.com/casszhao/temporal-drift-on-explanation}}
\end{abstract}

\section{Introduction}
One way of improving the transparency of deep learning models in natural language processing (NLP) is by extracting explanations that justify model predictions~\citep{lipton2018mythos,guidotti2018survey}. An explanation (i.e. rationale) consists of a subset of the input and is considered faithful when it accurately shows the reasoning behind a model's prediction \citep{zaidan-etal-2007-using, ribeiro-etal-2016-trust,deyoung2019eraser,jacovi-goldberg-2020-towards}. For example, removing a faithful rationale from the input should result into a prediction change. 
Two widely used methods for extracting rationales are (i) feature attribution methods that produce a distribution over the input tokens, indicating their contribution (i.e. importance) to the model's prediction \citep{ribeiro2016should, wiegreffe-pinter-2019-attention}; and (ii) select-then-predict models that consist of two components, a rationale extractor and a predictor. The rationale extractor extracts rationales, and the predictor is trained on extracted rationales so that its predictions are inherently faithful~\citep{lei-etal-2016-rationalizing,jain2020learning}.

Previous work has focused on evaluating explanation faithfulness in \textit{synchronous} settings where the training and testing data come from the same temporal distribution~\citep{serrano2019attention, jain-wallace-2019-attention, atanasova-etal-2020-diagnostic,guerreiro-martins-2021-spectra}, or out-of-domain settings~\citep{chrysostomou-aletras-2022-empirical} where the training and testing data come from a different domain regardless of temporal drifts in the testing data. However, human languages evolve \cite{weinreich1968empirical,kim-etal-2014-temporal,carrier2019because} as manifested by novel usages developed for existing words (e.g. \emph{mouse} is a mammal or a computer accessory) and new words and topics (e.g. \emph{covidiot} during the COVID-19 pandemic) that appear over time. Language evolution leads to temporal concept drifts and 
a diachronic degradation of model performance in many NLP tasks when these are evaluated in \textit{asynchronous} settings, i.e. training and testing data come from different time periods~\cite{jaidka-etal-2018-diachronic,agarwal2021temporal,lazaridou2021mind,sogaard-etal-2021-need,chalkidis-sogaard-2022-improved}. 

In this paper, for the first time, we extensively analyze the impact of temporal concept drift on model explanations. We evaluate the faithfulness of rationales extracted using eight feature attribution approaches and three select-then-predict models over six text classification tasks with chronological data splits. Our contributions are as follows:

\begin{itemize}

    \item We find that faithfulness is not consistent under temporal concept drift for rationales extracted with feature attribution methods (e.g. it decreases or increases depending on the method), with an attention-based method demonstrating the most robust faithfulness scores across datasets;
    \item We empirically show that select-then-predict models can be used in asynchronous settings when it achieves comparable performance to the full-text model;
    \item We demonstrate that sufficiency is not trustworthy evaluation metrics for explanation faithfulness, regardless of a synchronous or an asynchronous setting. 
    
\end{itemize}

\section{Related Work}

\subsection{Temporal Concept Drift in NLP}

Temporal model deterioration describes the \emph{true difference in system performance} when a system is evaluated on chronologically newer data~\citep{jaidka-etal-2018-diachronic,gorman-bedrick-2019-need}. This has been linked to changes in the data distribution, also known as \emph{concept drift} in early studies \cite{schlimmer1986incremental, widmer1993effective}. 
Previous work has demonstrated the impact of temporal concept drift on model performance by assessing the \emph{temporal generalization} \citep{lazaridou2021mind,sogaard-etal-2021-need,agarwal2021temporal,rottger-pierrehumbert-2021-temporal-adaptation}.
\citet{sogaard-etal-2021-need} has studied several factors that affect the true difference in system performance such as temporal drift, variations in text length and adversarial data distributions. They found that temporal variation is the most important factor for performance degradation and suggest including chronological data splits in model evaluation.
\citet{chalkidis-sogaard-2022-improved} also noted that evaluating on random splits with the same temporal distribution as the training data consistently over-estimates model performance at test time in multi-label classification problems.

Previous work on mitigating temporal concept drift includes automatically identifying semantic drift of words over time~\cite{tsakalidis-etal-2019-mining,giulianelli-etal-2020-analysing,rosin2022temporal,montariol-etal-2021-scalable}. Efforts have also been made to mitigate the impact of temporal concept drift on model prediction performance \citep{lukes-sogaard-2018-sentiment,rottger-pierrehumbert-2021-temporal-adaptation,loureiro-etal-2022-timelms,chalkidis-sogaard-2022-improved} and develop time-aware models \cite{10.1162/tacl_a_00459,rijhwani-preotiuc-pietro-2020-temporally,dhingra2021time,rosin2022temporal}. For example, both \citet{rottger-pierrehumbert-2021-temporal-adaptation} and \citet{loureiro-etal-2022-timelms} observed performance improvements when continue fine-tuning their models with chronologically newer data. 
While the impact of temporal concept drift on model performance has received particular attention, to the best of our knowledge, no previous work has examined its impact on model explanations.

\subsection{Concept Drift and Model Explanations}

\citet{poerner-etal-2018-evaluating} has compared the explanation quality between tasks that contain short and long textual context.
More recently, \citet{chrysostomou-aletras-2022-empirical} have studied model explanations in out-of-domain settings (i.e. under concept drift) using train and test data from different domains. Their results showed that the faithfulness of out-of-domain explanations unexpectedly increases, i.e. outperforming in-domain explanations' faithfulness. This is interesting given that performance degradation due to concept drift is often expected in domain adaptation \citep{schlimmer1986incremental, widmer1993effective, 10.3115/1220175.1220187, 10.1145/2523813}.

\section{Extracting Explanations}

We extract explanations using two standard approaches: (i) post-hoc methods; and (ii) select-then-predict models. 

\subsection{Post-hoc Explanation Methods}\label{sec:methodology_posthoc}

For post-hoc explanations, we fine-tune a BERT-base model on each task on the synchronous training set and extract explanations using post-hoc feature attribution methods for all synchronous and asynchronous testing sets.
We use eight widely used feature attribution methods following~\citet{chrysostomou-aletras-2021-enjoy,chrysostomou-aletras-2021-improving}. 

\begin{itemize}
\item{\bf Attention ($\alpha$):} Token importance is computed using the corresponding normalized attention scores \citep{jain2020learning}.

\item{\bf Scaled attention ($\alpha\nabla\alpha$)} Attention scores scaled by their corresponding gradients \citep{serrano2019attention}.

\item{\bf InputXGrad ($x\nabla x $)} Attributes importance by multiplying the input with its gradient computed with respect to the predicted class \citep{kindermans2016investigating, atanasova-etal-2020-diagnostic}.

\item{\bf Integrated Gradients (IG)} Ranks input tokens by computing the integral of the gradients taken along a straight path from a baseline input (zero embedding vector) to the original input \citep{pmlr-v70-sundararajan17a}. 

\item{\bf GradientSHAP (Gsp)} 
A gradient-based method to compute SHapley Additive exPlanations (SHAP) values for assigning token importance  \citep{lundberg2017unified}. Gsp computes the gradient of outputs with respect to randomly selected points between the inputs and a baseline distribution.

\item{\bf LIME} Ranks input tokens by learning a linear surrogate model using data points randomly sampled locally around the prediction \citep{ribeiro2016should}.

\item{\bf DeepLift (DL)} Computes token importance according to the difference between the activation of each neuron and a reference activation (i.e. zero embedding vector) \citep{pmlr-v70-shrikumar17a}.

\item{\bf DeepLiftSHAP (DLsp)} Similar to Gsp, DLsp computes the expected value of attributions based on DL across all input-baseline pairs, considering a baseline distribution \citep{lundberg2017unified}.
\end{itemize}

\subsection{Select-then-predict Models}

We also use three state-of-the-art select-then-predict models. Two are trained end-to-end \citep{bastings-etal-2019-interpretable,guerreiro-martins-2021-spectra} while the other one uses a feature attribution method as the rationale extractor \citep{jain2020learning} with a separate predictor component, trained on the extracted rationales.

\begin{itemize}
\item{\bf HardKUMA:}
\citet{bastings-etal-2019-interpretable} proposed a modified version of the end-to-end rationale extraction model introduced by \citet{lei-etal-2016-rationalizing}. Choosing rationales in a binary fashion by sampling from a Bernoulli distribution is replaced with a Kumaraswamy distribution \citep{kumaraswamy1980generalized} to support continuous random variables. This way, the model is differentiable and easier to train. 

\item{\bf SPECTRA:} HardKUMA provides stochastic rationales due to the marginalization over all possible rationales and the sampling process. \citet{guerreiro-martins-2021-spectra} proposed SPECTRA, a model that uses LP-SparseMAP \citep{niculae2020lp} to obtain a deterministic rationale extraction process. \citet{niculae2020lp} have experimented with three different factor graphs showing that XorAtMostOne outperforms the other two (i.e. Budget, AtMostOne2). We use SPECTRA with XorAtMostOne in our experiments. 
For HardKUMA and SPECTRA, we use a Bi-LSTM \citep{hochreiter1997long} because it has been shown to outperform BERT-based models \citep{guerreiro-martins-2021-spectra}.

\item{\bf FRESH:} \citet{jain2020learning} proposed FRESH, a model that first extracts rationales from a trained model (e.g. using a feature attribution method) and subsequently trains a classifier on the extracted rationales. We extract the top 20\% rationales using $\alpha \nabla \alpha$ that achieved the best performance in early experimentation. We also use BERT-base for the extraction and predictor components following \citet{jain2020learning}.

\end{itemize}

\section{Experimental Setup}

\subsection{Tasks and Data}

\renewcommand{\arraystretch}{1.3}
\begin{figure}[!t]
    \centering
    \includegraphics[width=.23\textwidth]{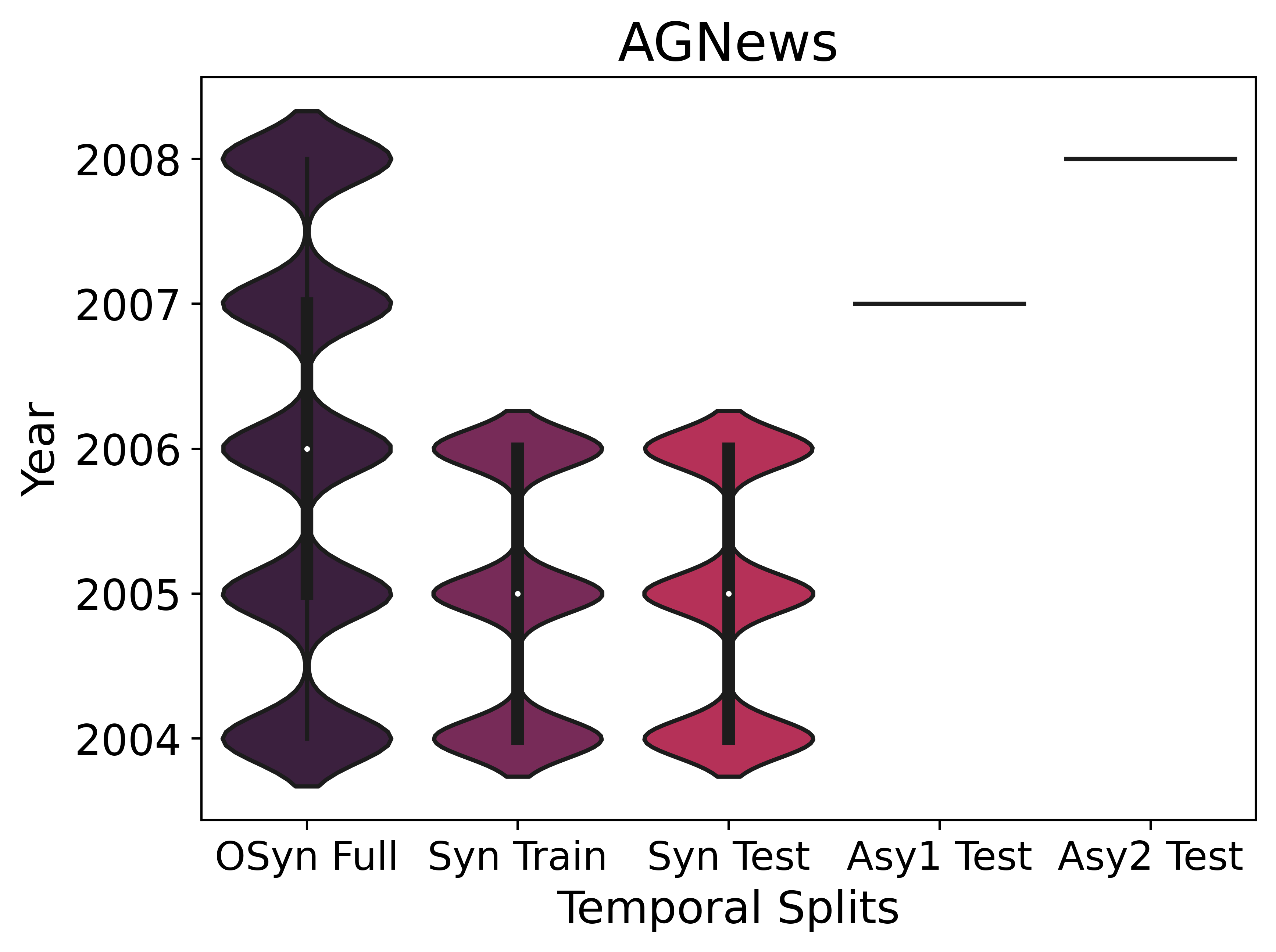}
    \includegraphics[width=.23\textwidth]{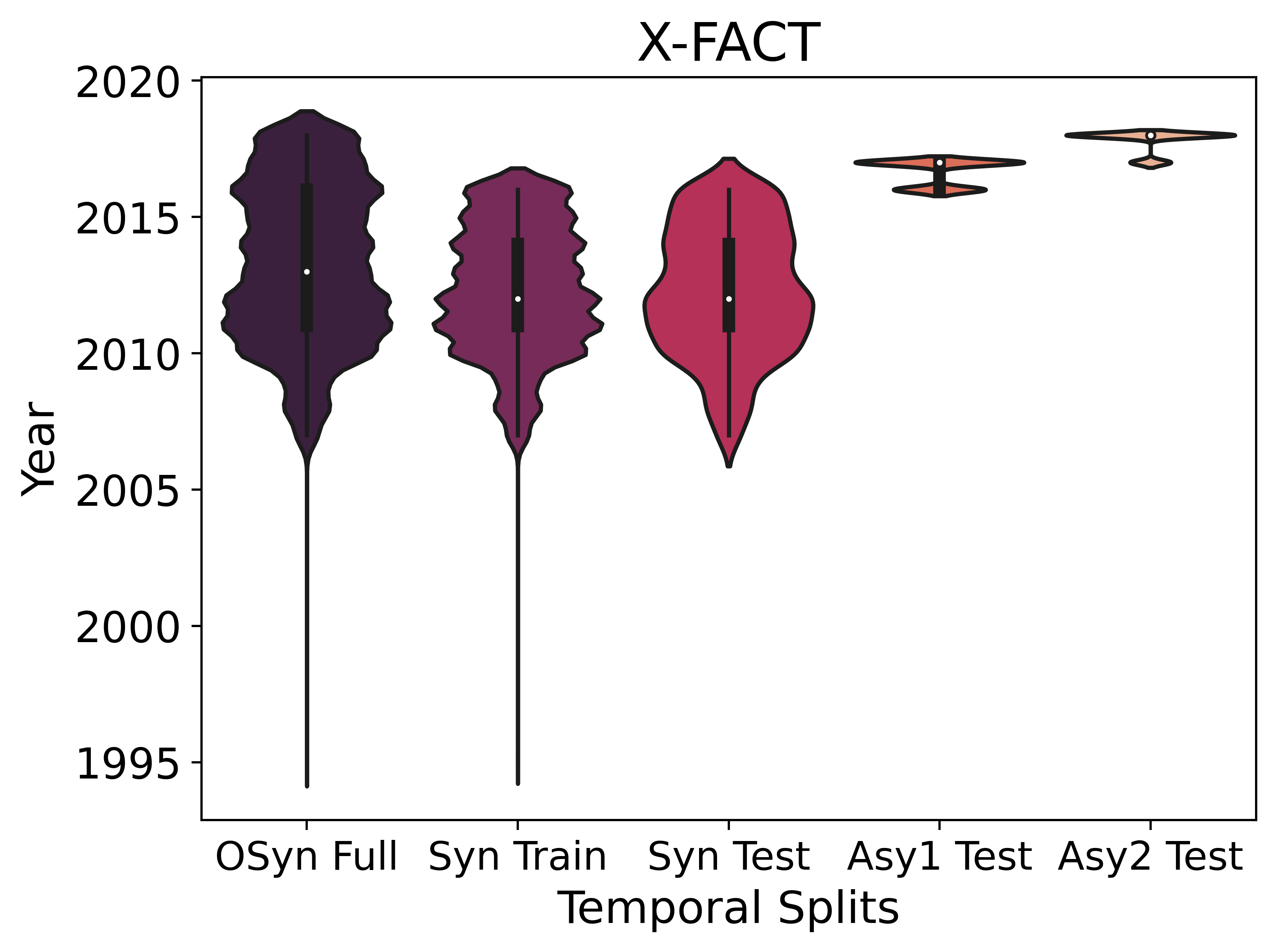}\\
    \includegraphics[width=.23\textwidth]{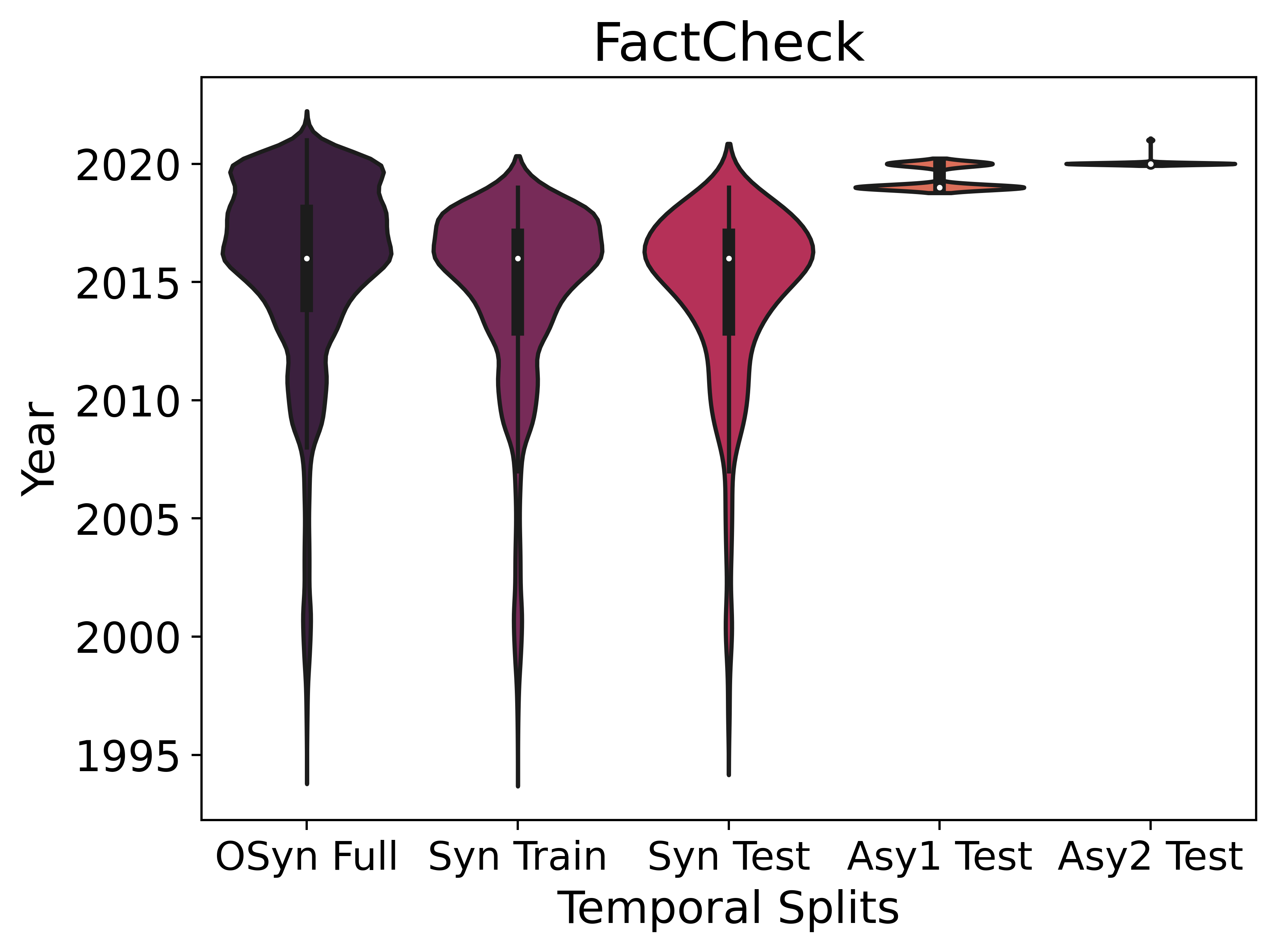}
    \includegraphics[width=.23\textwidth]{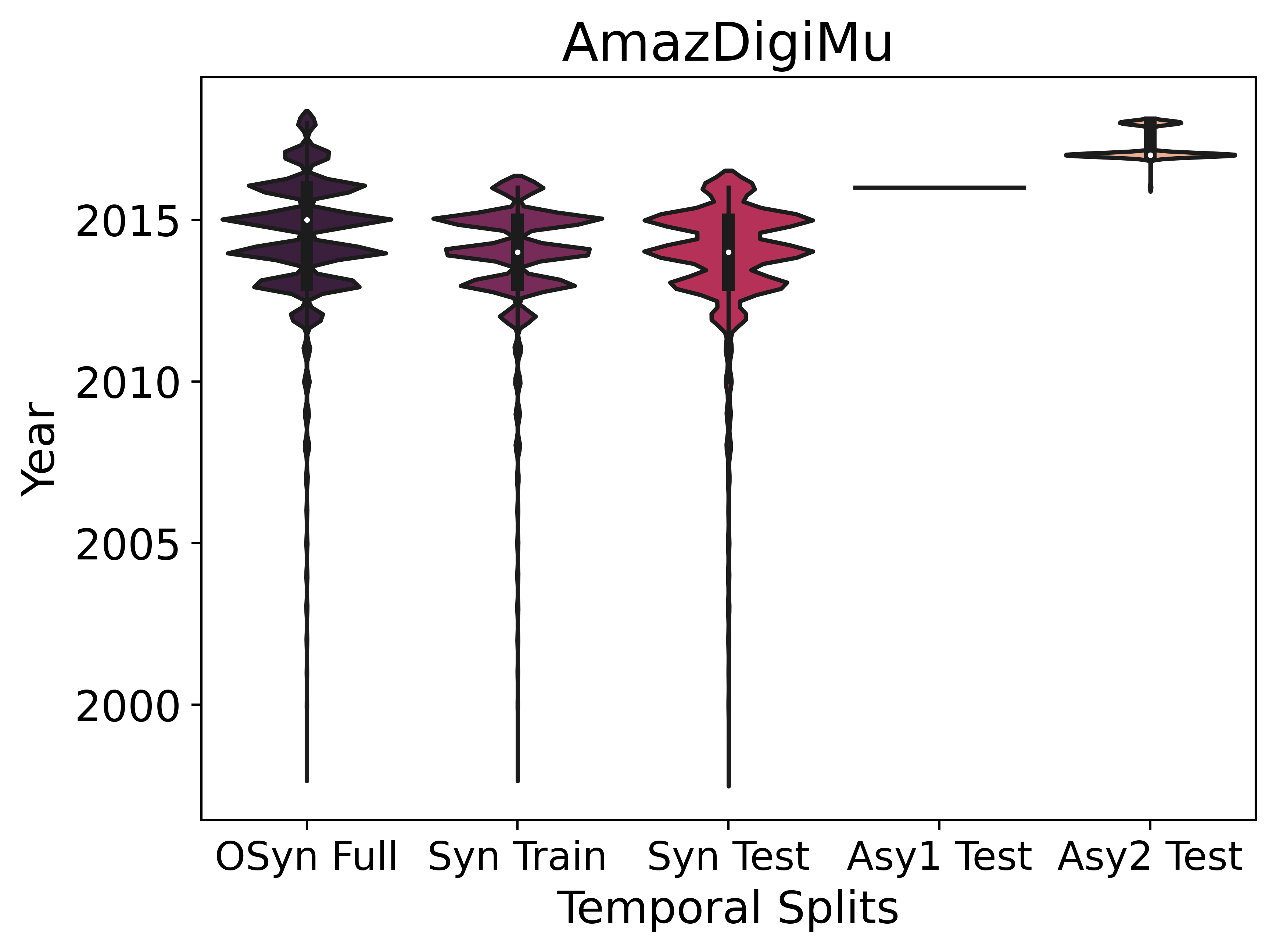}\\
    \includegraphics[width=.23\textwidth]{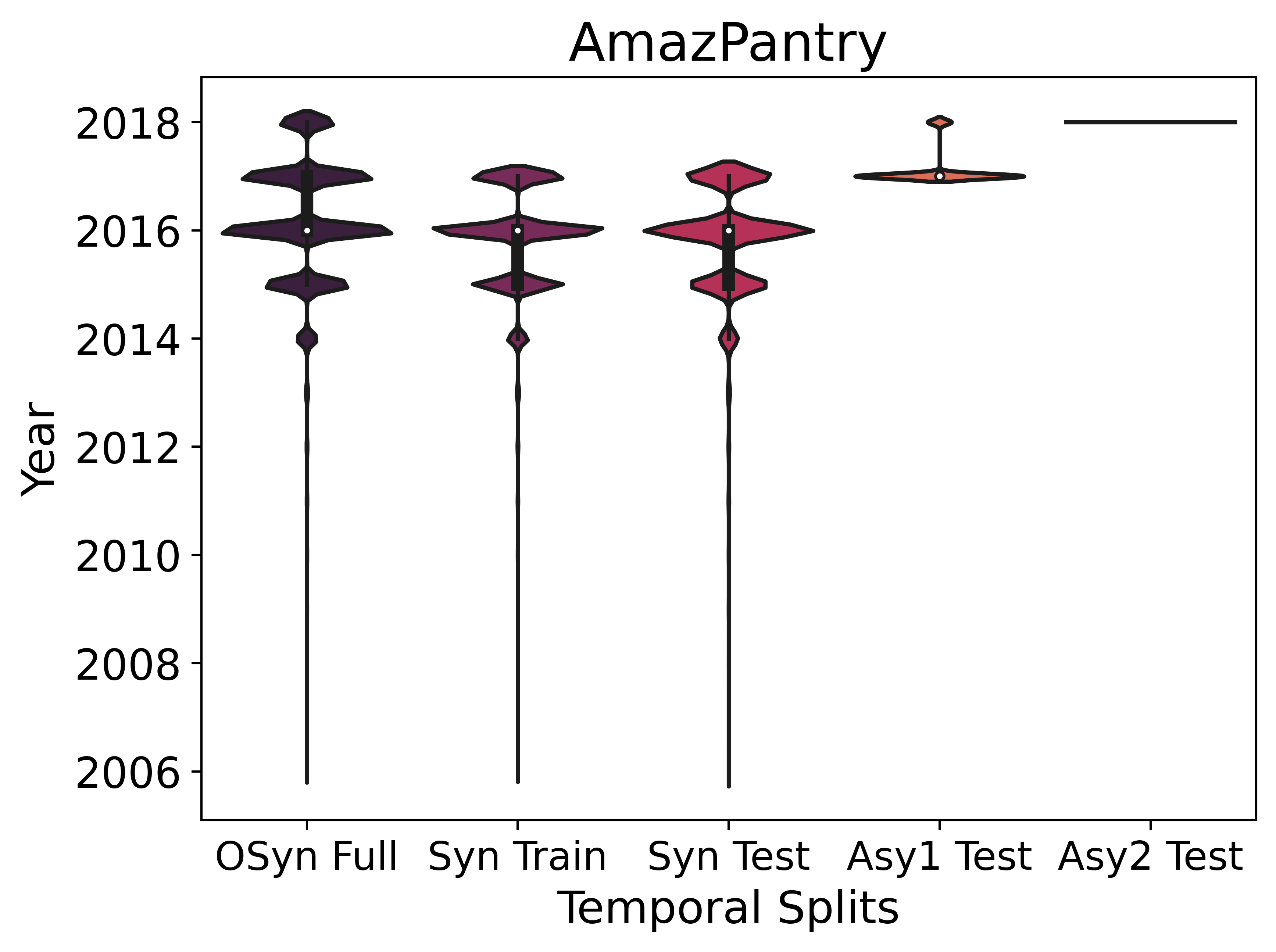}
    \includegraphics[width=.22\textwidth]{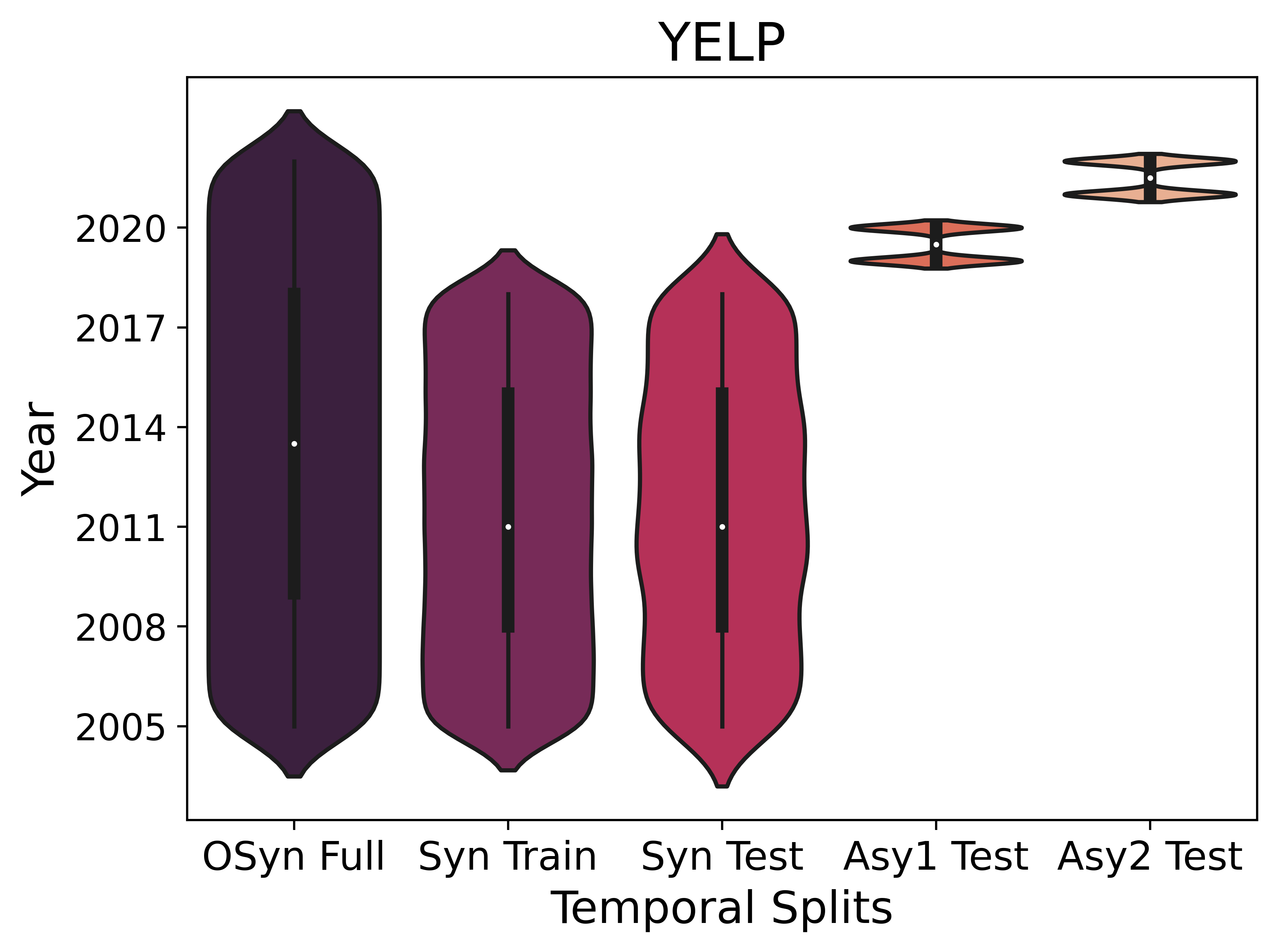}
\caption{Density curves for time distribution across temporal splits and the original full size dataset for each task.}
\label{fig:time_distribution}
\end{figure}

\renewcommand{\arraystretch}{1.2}
\begin{table}[!t]
\resizebox{\linewidth}{!}{%
\begin{tabular}{@{}p{0.23\linewidth}p{0.15\linewidth}lrrp{0.17\linewidth}p{0.17\linewidth}@{}}
\toprule
{\bf Task} & {\bf \#Classes} & {\bf Splits} & \multicolumn{1}{l}{\bf Start Date} & \multicolumn{1}{l}{\bf End Date} & {\bf Span} (Days) & {\bf \#Data}\\ \midrule

\multirow{4}{*}{AGNews} & \multirow{4}{*}{4} 
  & Train & 2004-08-18 & 2006-12-20 & 854 & 9358 \\
 &  & Syn Test & 2004-08-18 & 2006-12-20 & 854 & 9358 \\
 &  & Asy1 Test & 2007-01-30 & 2007-12-31 & 335 & 9358 \\
 &  & Asy2 Test & 2008-01-01 & 2008-02-20 & 50 & 9358 \\ \midrule

\multirow{4}{*}{X-FACT} & \multirow{4}{*}{6} 
  & Train & 1995-04-01 & 2016-08-31 & 7823 & 7232 \\
 &  & Syn Test & 2007-01-04 & 2016-08-31 & 3527 & 1204 \\
 &  & Asy1 Test & 2016-08-31 & 2017-09-30 & 395 & 1205 \\
 &  & Asy2 Test & 2017-09-30 & 2018-11-12 & 408 & 1204 \\ \midrule
 
\multirow{4}{*}{FactCheck} & \multirow{4}{*}{2} 
 & Train & 1995-09-25 & 2019-05-01 & 8619 & 7446 \\
 &  & Syn Test & 1996-08-02 & 2019-05-01 & 8307 & 1241 \\
 &  & Asy1 Test & 2019-05-02 & 2020-05-15 & 379 & 1368 \\
 &  & Asy2 Test & 2020-05-15 & 2021-07-19 & 430 & 1368 \\ \midrule
 
\multirow{4}{*}{AmazDigiMu} & \multirow{4}{*}{3} 
 & Train & 1998-08-21 & 2016-05-07 & 6469 & 101774 \\
 &  & Syn Test & 1998-12-20 & 2016-05-07 & 6351 & 16963 \\
 &  & Asy1 Test & 2016-05-07 & 2016-12-30 & 237 & 16962 \\
 &  & Asy2 Test & 2016-12-30 & 2018-09-26 & 635 & 16962 \\ \midrule
 
\multirow{4}{*}{AmazPantry} & \multirow{4}{*}{3} 
  & Train & 2006-04-28 & 2017-07-30 & 4111 & 82566 \\
 &  & Syn Test & 2006-12-22 & 2017-07-30 & 3873 & 13762 \\
 &  & Asy1 Test & 2017-07-30 & 2018-01-21 & 175 & 13761 \\
 &  & Asy2 Test & 2018-01-21 & 2018-10-04 & 256 & 13761 \\ \midrule
 
 \multirow{4}{*}{Yelp} & \multirow{4}{*}{5} 
  & Train & 2005-02-16 & 2018-12-31 & 5066 & 8540 \\
 &  & Syn Test & 2005-02-16 & 2018-12-24 & 5059 & 1708 \\
 &  & Asy1 Test & 2019-01-01 & 2020-12-31 & 730 & 1708 \\
 &  & Asy2 Test & 2021-01-01 & 2022-01-19 & 383 & 1708 \\ \bottomrule
\end{tabular}%
}
\caption{Data statistics and the temporal splits for each task.}
\label{tab:data_stat}
\end{table}

\paragraph{Tasks}
We evaluate all methods on three diverse text classification tasks including six different datasets: (1) topic classification; (2) misinformation detection; and (3) sentiment analysis:

\begin{itemize}
    \item {\bf AGNews:} Topic classification across four topics (Business, Sports, Science/Technology and World) from AG News~\citep{10.1145/1060745.1060764};

    \item {\bf X-FACT:} Factual correctness classification of short statements into five classes \citep{gupta-srikumar-2021-x}: True, Mostly-True, Partly-True, Mostly-False and False;

    \item {\bf FactCheck:} Binary classification of potential misinformation stories as truthful or misinformation \citep{jiang-wilson-2021-structurizing};

    \item {\bf Amazon Reviews:} We predict the sentiment (negative, neutral, positive) of Amazon product reviews from digital music (\textbf{AmazDigiMu}) and pantry (\textbf{AmazPantry}) as \citet{ni-etal-2019-justifying};

    \item {Yelp:} Multi-class sentiment classification (positive, negative) following \citet{Zhang2015-charcnn}.
\end{itemize}

\paragraph{Data Splits}
To simulate temporal concept drifts, we create different chronological splits according to the time-stamps of the data points in each dataset. We split each dataset into a training set and three different test sets. The time spans of the three test sets follow a chronological order without any overlapping. The test set with the earliest time span (\emph{Syn}) has the exact same time span as the training data (i.e. a synchronous setting). The other two splits denoted as \emph{Asy1} and \emph{Asy2} that are chronologically newer correspond to asynchronous settings. 
Figure \ref{fig:time_distribution} shows the temporal distribution of each data split compared to the original data. Table \ref{tab:data_stat} summarizes the key statistics for each split. More details for the data and tasks can be found in the Appendix \ref{app:data_stat}. We also provide results of all models on the original (synchronous) test set (\emph{OSyn}).

\begin{figure*}[t!] 
\centering
    \includegraphics[width=0.85\textwidth]{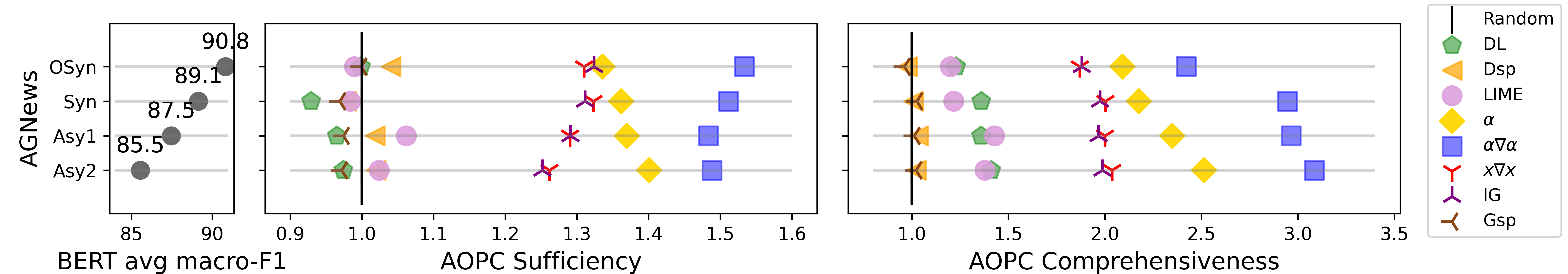}
    \includegraphics[width=0.85\textwidth]{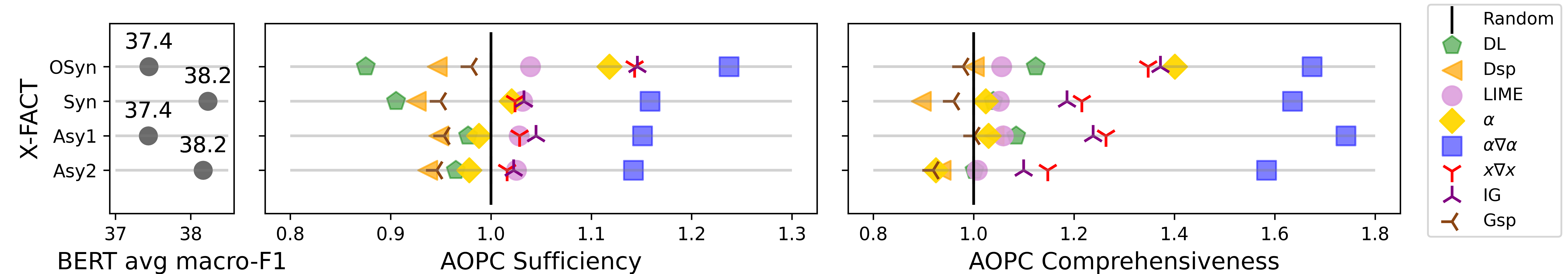}
    \includegraphics[width=0.85\textwidth]{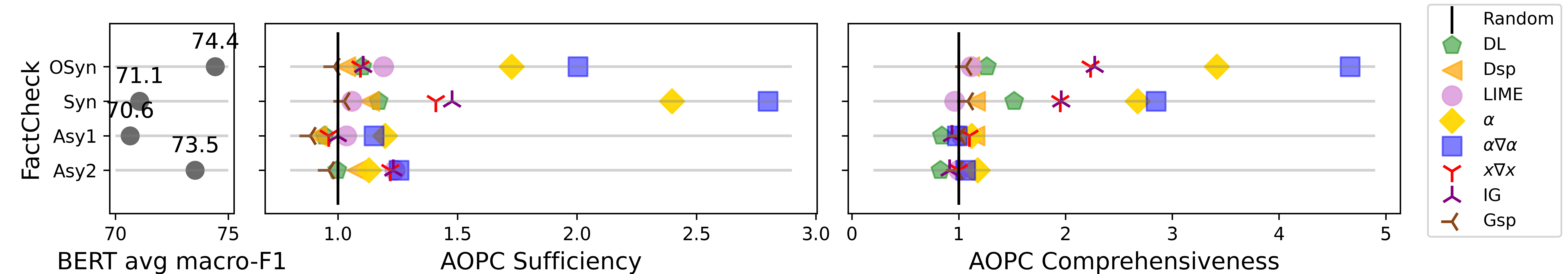}
    \includegraphics[width=0.85\textwidth]{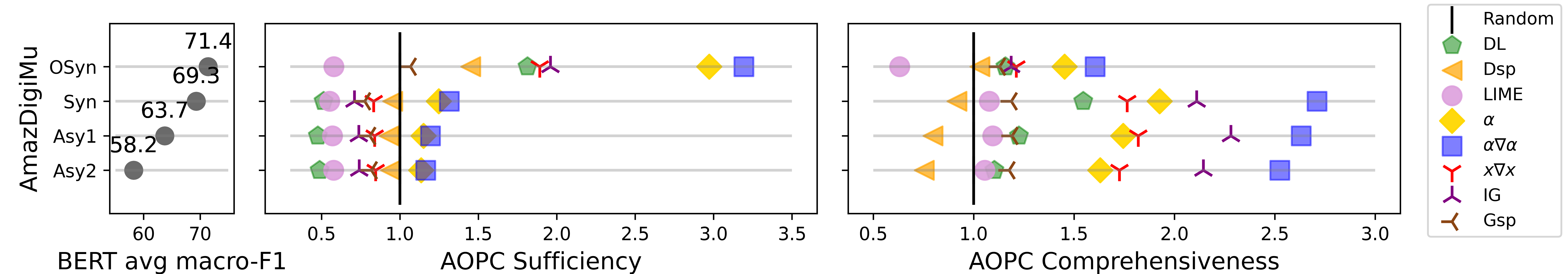}
    \includegraphics[width=0.85\textwidth]{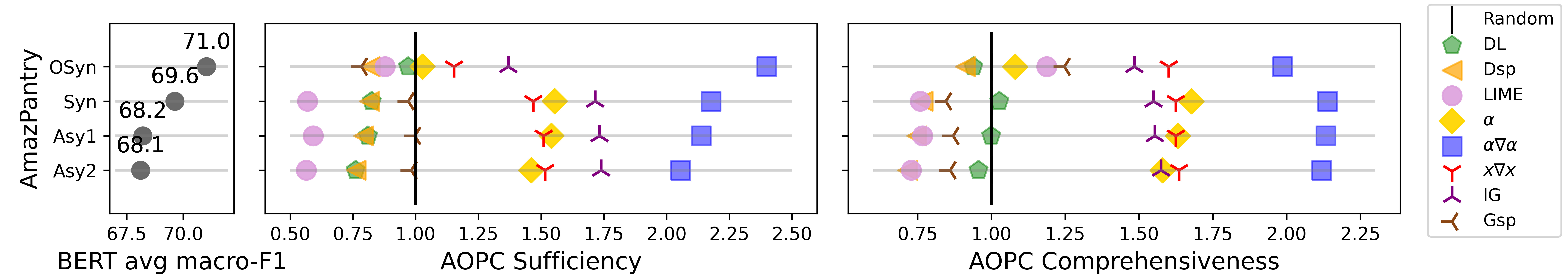}
    \includegraphics[width=0.85\textwidth]{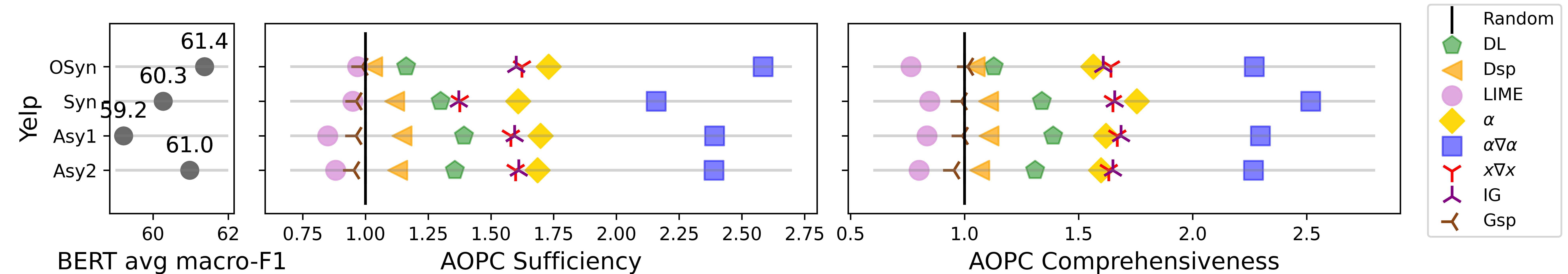}
	\caption{Post-hoc AOPC normalized sufficiency and comprehensiveness (higher is better) of the original test set (OSyn), our Syn and Asy splits, across feature attribution approaches and a random baseline.}\vspace*{0pt}
	\label{fig:posthoc_resutls_topk}
\end{figure*}

\subsection{Evaluation}
For each task, we train a model on the training set and then evaluate post-hoc explanations and select-then-predict performance on our three chronological splits, namely \emph{Syn}, \emph{Asy1} and \emph{Asy2}.

\paragraph{Post-hoc Explanations}\label{sec:Post-hoc_explanations}
We evaluate the faithfulness of post-hoc explanations using two popular metrics \citep{deyoung2019eraser,carton-etal-2020-evaluating}:
\begin{itemize}
    \item \textbf{Normalized Sufficiency} quantifies how sufficient a rationale is for making the same prediction $p(\hat{y}|\mathcal{R})$ to the prediction of the full text model $p(\hat{y}| \mathbf{x})$. We use the normalized version to allow a fairer comparison across models and tasks:
\begin{equation}
\small
\begin{aligned}
    \text{Suff}(\mathbf{x}, \hat{y}, \mathcal{R}) = 1 - max(0, p(\hat{y}| \mathbf{x})- p (\hat{y}|\mathcal{R})) \\
    \text{NormSuff}(\mathbf{x}, \hat{y}, \mathcal{R}) = \frac{\text{Suff}(\mathbf{x}, \hat{y}, \mathcal{R}) - \text{Suff}(\mathbf{x}, \hat{y}, 0)}{1 - \text{Suff}(\mathbf{x}, \hat{y}, 0)}
 \end{aligned}
\end{equation}

    \item \textbf{Normalized Comprehensiveness} assesses how much information the rationale holds, measuring changes in predictions when masking the rationale $p(\hat{y}|\mathbf{x}_{\backslash\mathcal{R}})$. Similar to sufficiency, we use the normalized version: 
\begin{equation}
    \small
    \begin{aligned}
        \text{Comp}(\mathbf{x}, \hat{y}, \mathcal{R}) = max(0, p(\hat{y}| \mathbf{x})- p (\hat{y}|\mathbf{x}_{\backslash\mathcal{R}})) \\
        \text{NormComp}(\mathbf{x}, \hat{y}, \mathcal{R}) = \frac{\text{Comp}(\mathbf{x}, \hat{y}, \mathcal{R})}{1 - \text{Suff}(\mathbf{x}, \hat{y}, 0)}
    \end{aligned}
    \end{equation}
\end{itemize}

Further, we evaluate explanations of different lengths (top 2\%, 10\%, 20\% and 50\% of tokens extracted) and report the ``Area Over the Perturbation Curve'' (AOPC) of Normalized Sufficiency and Normalized Comprehensiveness following \citet{deyoung2019eraser}. 

Following \citet{chrysostomou-aletras-2022-empirical}, a baseline random attribution (i.e. randomly assigning importance scores) is used as a yardstick to allow a comparison across chronological splits and tasks. We use the ratio between the comprehensiveness or sufficiency score of a feature attribution and the score of the random baseline to compute its final faithfulness score. Faithfulness scores under 1.0 indicate that the rationale for a particular feature attribution is less faithful than just randomly selecting input tokens as a rationale. 

We avoid using metrics such as RemOve And Retrain (ROAR) due to their demanding computation requirements \citep{hooker2019benchmark}. We also omit the use of other popular metrics such as Word Relevance \citep{arras-etal-2019-evaluating} and Fraction of Tokens \citep{serrano-smith-2019-attention} as they are similar to comprehensiveness and sufficiency.

\paragraph{Select-then-predict}

Select-then-predict classifiers are trained only on rationales and discard the rest of the input, hence they are inherently faithful. As such, one way to check how good their extracted explanations are, is to compare their predictive performance to the full-text trained model following \citet{jain2020learning}. A good rationale should achieve high predictive performance retention compared to the full-text model.
We, therefore, compare the predictive performance of the three select-then-predict models with corresponding models trained on full-text: (1) FRESH against a BERT-base model; and (2) HardKUMA and SPECTRA against a Bi-LSTM with the same number of layers, pre-trained embeddings and hidden dimensions.


\section{Results}\label{sec:results}

\subsection{Post-hoc Explanation Faithfulness}

We hypothesize that when the predictive performance of a model drops in asynchronous settings, the sufficiency and faithfulness scores that are based on predictive likelihood should also drop. Our hypothesis is based on the assumption that the lower the predictive performance, the lower the predictive likelihood for a well-calibrated model~\citep{desai2020calibration}.

Figure \ref{fig:posthoc_resutls_topk} shows the AOPC normalized sufficiency and comprehensiveness scores for each feature attribution method across temporal splits, with the corresponding model predictive performance along the left side. Full results can be found in Table \ref{tab:predictive_and_faithfulness} in the Appendix.

We first observe that certain feature attributions (e.g. $x\nabla x$, $\alpha\nabla \alpha$, $IG$) score above the random baseline in the majority of settings, suggesting that they remain faithful in asynchronous settings. The attention-based $\alpha\nabla \alpha$ in particular, outperforms not only the random baseline by a large margin in all settings, but also the rest of the feature attributions tested in the majority of cases. 
For example, across all temporal splits in Yelp, $\alpha\nabla \alpha$ scores higher than 2.16 in both sufficiency and comprehensiveness, i.e. compared to the random baseline. The second-best one,  $\alpha$, is again an attention-based attribution method and only scores 1.60x better. Two other feature attribution methods (LIME and Gsp) fail to score above the random baseline.
On the other hand, certain methods such as DL, also fail to outperform the random baseline in general. This suggests that they cannot be trusted in asynchronous settings. For instance, Gsp fails to exceed the random baseline across all tasks for sufficiency scores on Asy splits while DL only scores slightly higher than random baseline (1.36 and 1.37 respectively).

Contrary to our initial hypothesis, \emph{the faithfulness scores of all feature attribution methods do not necessarily fluctuate together with predictive performance, when comparing between synchronous and asynchronous settings}.

For example, in AGNews, AmazDigiMu and AmazPantry, predictive performance decreases along with chronological order. However, looking into sufficiency, only $x\nabla x$ in AGNews, $\alpha\nabla \alpha$  and $\alpha$ in both AmazDigiMu and AmazPantry confirm our initial hypothesis, i.e. the faithfulness decreases along with predictive performance. In the rest of the cases for these three tasks, we do not observe any pattern between faithfulness and the chronological order of the test data.

\subsection{Select-then-predict Predictive Performance}

\begin{figure}[!t] 
\centering
    \includegraphics[width=0.9\columnwidth]{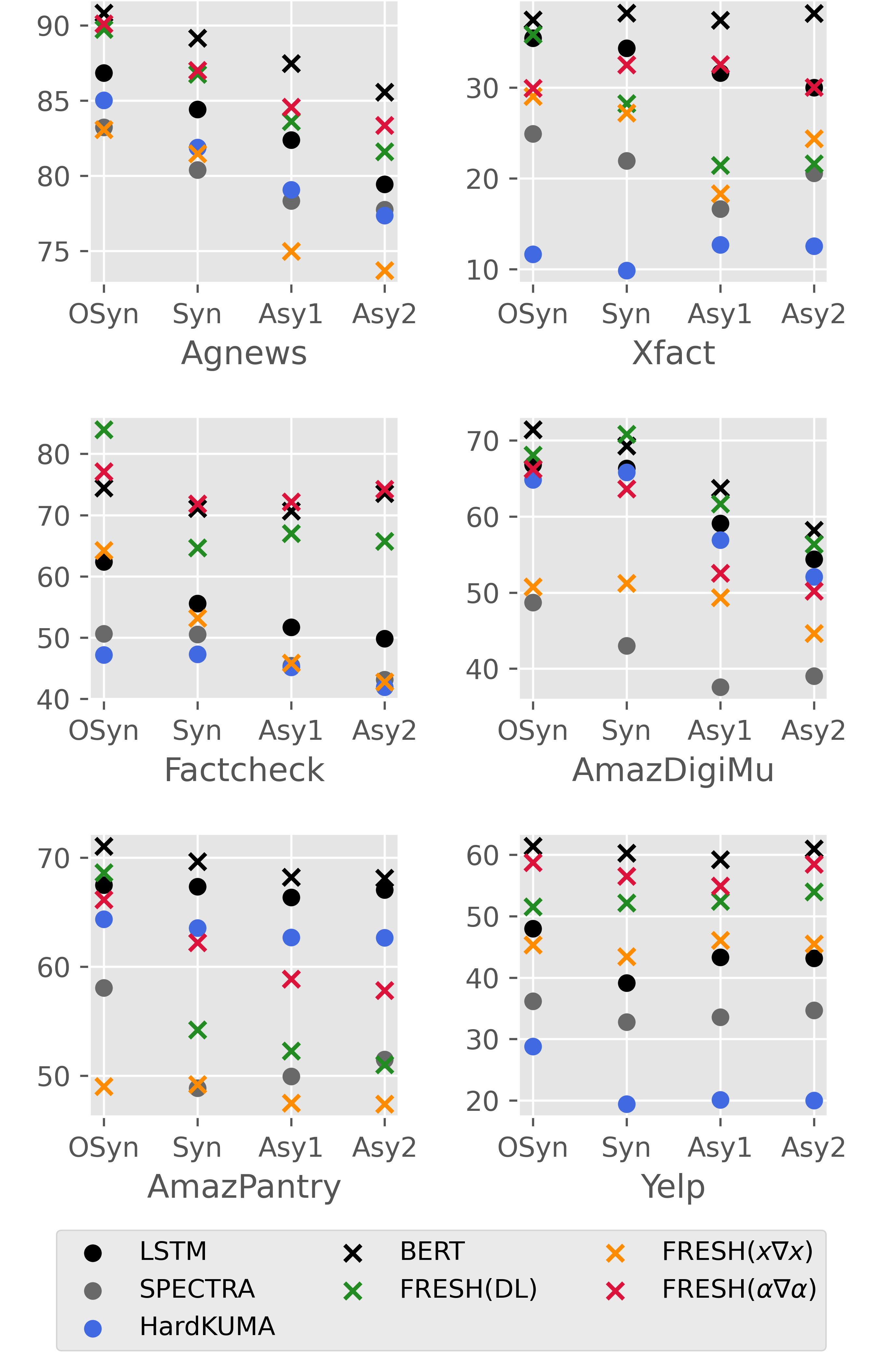}	\caption{Averaged macro F1 performance (5 runs) of select-then-predict methods and models trained on full-text.}\vspace*{0pt}
	\label{fig:selective}
\end{figure}

Figure \ref{fig:selective} shows the macro F1 scores (i.e. averaged over 5 runs with different random seeds) for the three select-then-predict models and their full-text trained counterparts. We compare full-text trained BERT against FRESH using the most faithful feature attribution ($\alpha\nabla\alpha$), one that is close to the average faithfulness of all methods ($x\nabla x $) and the least faithful one (DL). 
We also compare HardKUMA and SPECTRA against a full-text trained Bi-LSTM. For the full stack of results (including standard deviations), see Appendix \ref{app:sec:select_then_predict}.

\begin{figure}[!t]
    \centering
    \begin{subfigure}[b]{0.35\textwidth}
    \includegraphics[width=\textwidth]{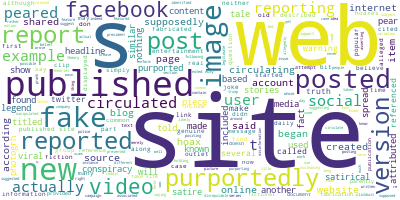}
    \caption{FactCheck (Syn)}
    \end{subfigure}
    \vspace{0.0em}
    \begin{subfigure}[b]{0.35\textwidth}
    \includegraphics[width=\textwidth]{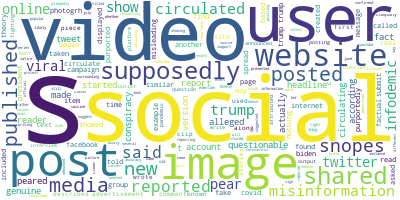}
    \caption{FactCheck (Asy2)}
    \end{subfigure}
    \vspace{0.0em}
    \begin{subfigure}[b]{0.35\textwidth}
    \includegraphics[width=\textwidth]{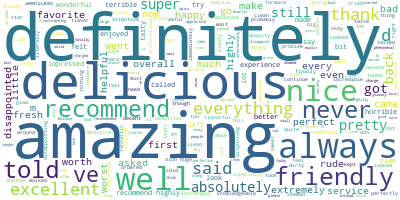}
    \caption{Yelp (Syn)}
    \end{subfigure}
    \vspace{0.0em}
    \begin{subfigure}[b]{0.35\textwidth}
    \includegraphics[width=\textwidth]{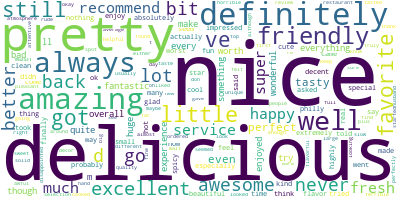}
    \caption{Yelp (Asy2)}
    \end{subfigure}
\caption{Wordclouds from synchronous (Syn) and asynchronous (Asy2) test sets for FactCheck and Yelp.}
\label{fig:wordclouds}
\end{figure}

\paragraph{HardKUMA \& SPECTRA}

As expected, the two models result in predictive performance drops compared to the full-text trained Bi-LSTM in the synchronous data splits (i.e. OSy and Syn). 
In asynchronous settings (i.e. Asy1 and Asy2), there is no consistent pattern observed. In certain tasks, their predictive performance increases (e.g. SPECTRA in AmazPantry in both Asy1 and Asy2) while in other tasks decreases (e.g. for SPECTRA and HardKUMA in AGNews). For example, the performance of HardKUMA drops gradually from 85\% for OSyn to 77\% in Asy2. Similarly, the performance for SPECTRA drops from 50\% to approximately 40\% in FactCheck.

An interesting observation is that the predictive performance of HardKUMA and SPECTRA is comparable to the model trained on full-text in asynchronous settings in cases the two models have also achieved a comparable performance in synchronous settings. For example, HardKUMA exhibits comparable performance across all settings with the full-text trained Bi-LSTM in AmazDigiMu. 

\textit{We therefore suggest, that HardKUMA and SPECTRA are reliable in asynchronous settings, when only their performance is comparable to the full-text model in synchronous settings.}

\paragraph{FRESH}
We hypothesize that FRESH trained with rationales extracted from a faithful feature attribution method (i.e. its sufficiency is substantially higher relative to the random baseline), it should result into comparable predictive performance of the full-text trained model.
In theory, these rationales should contain `sufficient' information for a classifier to perform comparably to the full-text trained model.

We first observe that FRESH with $\alpha\nabla\alpha$ generally mirrors BERT's performance across all settings in most tasks, with the only exception in X-FACT (see Figure~\ref{fig:selective}). We speculate that a possible reason for this, is the larger number of classes together with the small size of the dataset.
This behavior also indicates that FRESH using the most faithful attention-based attribution method, $\alpha\nabla\alpha$ is not impacted by temporal drifts, more than its full-text trained counterpart. For example, the performance of FRESH($\alpha\nabla\alpha$) in X-FACT, FactCheck remains mostly stable across different test splits.
In comparison, we do not observe the same mirroring behavior of FRESH train with less sufficient rationales from attribution methods such as DL and $x\nabla x $, across splits and tasks. 

Comparing between faithfulness scores (sufficiency and comprehensiveness) and FRESH predictive performance, we identify a counter-intuitive pattern. In sharp contrast to our initial expectations, using rationales extracted from the lowest scoring feature attribution for sufficiency (i.e. DL), results in higher predictive performance for FRESH compared to the more sufficient rationales extracted with $x\nabla x$. For example, in AGNews, DL consistently scores below the random baseline for sufficiency in all settings, whilst $x\nabla x$ remains consistently more sufficient, scoring above the random baseline (see Figure \ref{fig:posthoc_resutls_topk}). However, the performance of FRESH trained on rationales extracted with DL is directly comparable to using $\alpha\nabla\alpha$ for rationale extraction across all settings. 
Its performance is also closer to the full-text trained model. Using $x\nabla x$ rationales to train FRESH, it results into lower predictive performance compared to FRESH(DL). 
We further investigate these conflicting patterns in Section \ref{sec:analysis_fresh}.

To summarize, \emph{FRESH trained on rationales extracted by a robust and faithful feature attribution (i.e. $\alpha\nabla\alpha$) is reliable in asynchronous settings and not impacted by temporal drifts compared to the full-text model. FRESH trained with less faithful rationales, such as DL and $x\nabla x $, is not reliable across tasks}.

\section{Qualitative Analysis}

We also conduct a qualitative analysis on the rationales extracted by $\alpha\nabla\alpha$, to find possible reasons that justify its stability and robustness when moving from synchronous to asynchronous settings (i.e. invariance to concept drift), when using FRESH. Figure \ref{fig:wordclouds} shows wordclouds (larger words appear more frequently in the rationales) for FactCheck (a, b) and Yelp (c, d), on the synchronous split Syn and the asynchronous split Asy2.

Starting with FactCheck, we observe that salient words change when moving to asynchronous settings. For example, in Syn, the extracted rationales contain words like ``web'', ``site'' and ``published''. In contrast, rationales from Asy2 contain words like ``video'', ``social'' and ``post''. These indicate a shift in how misinformation is spread across time (i.e. different types of media), which surprisingly is picked up by the rationales when moving to asynchronous settings. Similarly in Yelp, whilst the majority of most frequent words remains similar across chronological splits (e.g. ``delicious'', ``nice'', ``amazing''), we still observe some concept drift that is again picked up by the rationales. For example, it appears that more recent restaurant reviews are concerned about the experience and appearance of the restaurant, as picked up in Asy2. This is highlighted by the fact that Asy2 contains frequently the words ``experience'' and ''pretty'' (to describe the place and food, we found several examples through a manual analysis that refer to these two concepts). On the other hand, we note that Syn does not contain words relevant to the experience and appearance of the place.

\section{To Trust Sufficiency or Not?}\label{sec:analysis_fresh}

The contradictory patterns observed between post-hoc explanations and FRESH (see Section \ref{sec:results}), questions the efficacy of using sufficiency to measure faithfulness in asynchronous settings. Inspired by the explanation-game \citep{treviso-martins-2020-explanation}, we use the classifier from FRESH as a layperson and measure its ability to generate the same predictions as the full-text trained model. Our hypothesis is that if a feature attribution produces highly sufficient rationales, the layperson should also have a high predictive performance (when using the full-text model's predictions as gold labels) and vice versa. If sufficiency is reliable as a metric, we therefore expect that the most sufficient rationales to be obtained using $\alpha\nabla\alpha$, followed by $x\nabla x$ and the least sufficient to be DL.

\begin{figure}[!t] 
\centering
    \includegraphics[width=0.9\columnwidth]{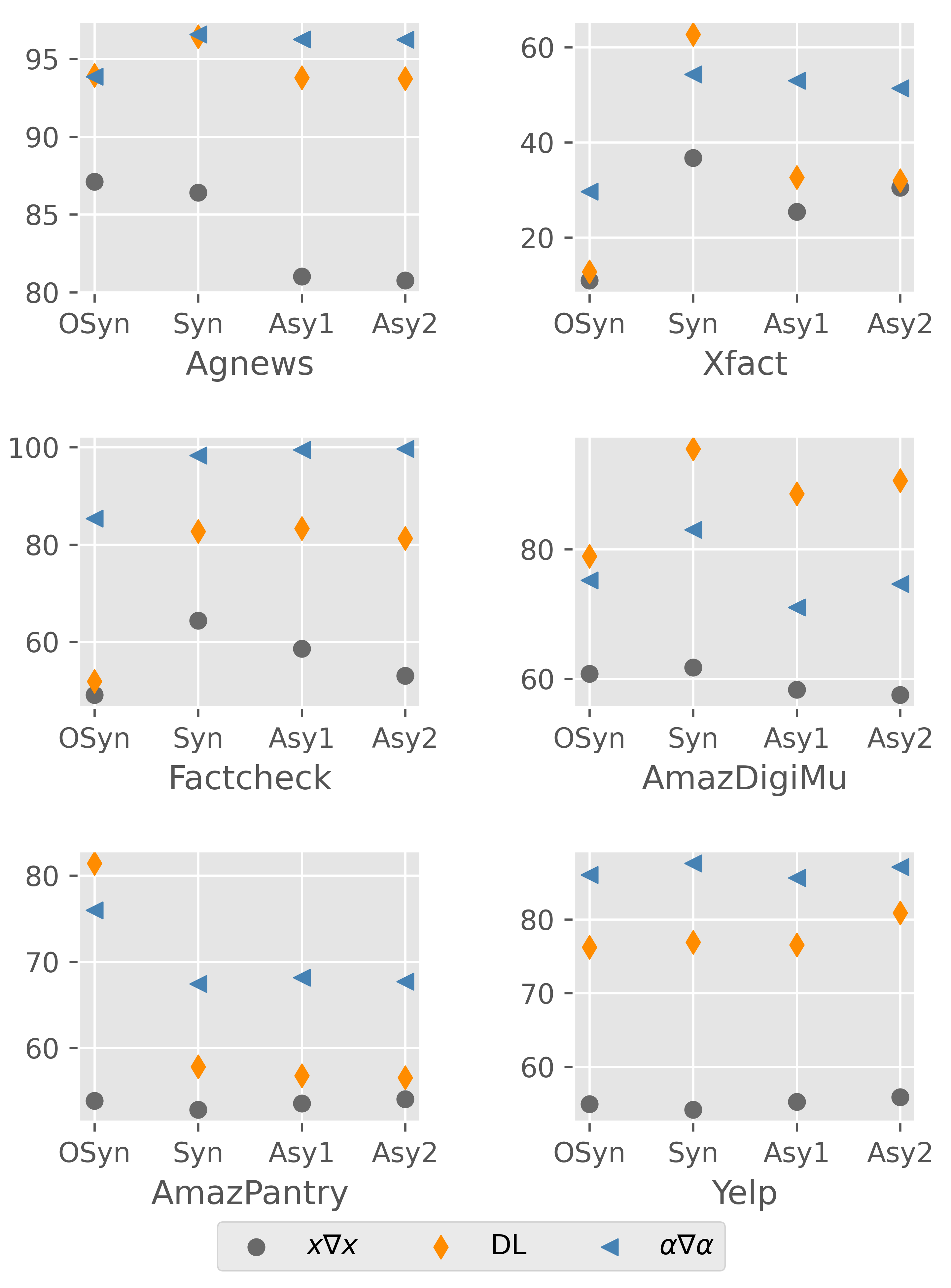}
	\caption{Averaged macro F1 performance of FRESH trained on $\alpha\nabla\alpha$, DL and $x\nabla x $ to predict  the labels predicted by a full-text model (BERT). .}
\label{fig:fresh_compare_bert_accurate}
\end{figure}

Figure \ref{fig:fresh_compare_bert_accurate} shows the performance of the layperson (i.e. FRESH), in predicting the original predictions of the full-text trained model (i.e. a higher score denotes a higher agreement). We first observe that $\alpha\nabla\alpha$ outperforms in most datasets both DL and $x\nabla x$. For example, in FactCheck, $\alpha\nabla\alpha$ shows almost perfect agreement (approximately 100\%) with the full-text trained model in Syn and both Asy splits, highlighting the efficacy of this feature attribution in extracting faithful explanations. 

Contrary to our expectations and similar to observations with FRESH, DL consistently outperforms $x\nabla x$, even outperforming $\alpha\nabla\alpha$ in certain cases. For example in Yelp across both asynchronous settings, using DL the layperson is able to reach the same predictions as the full-text model, in approximately 80\% of the instances. In comparison, using $x\nabla x$, the layperson reaches the same predictions in only 55\% of the instances. 

Our findings suggest that \emph{sufficiency, as a metric for measuring faithfulness, cannot be trusted in asynchronous settings and also raises concerns for synchronous settings.}

\section{Conclusion}
We conducted an extensive empirical study to shed light on the impact of temporal drift on model explanations in asynchronous settings, including post-hoc methods and select-then-predict models. We demonstrate that faithfulness is not consistent under temporal variations across feature attribution methods, while select-then-predict models are mostly robust with negligible drops in predictive performance. In the future, we plan to extend our study into more tasks and data from different languages. We also plan to explore whether instance specific feature attribution improves faithfulness in asynchronous settings~\citep{chrysostomou2022flexible}.

\section*{Limitations}
This study focuses only on experimenting with data in English. We would expect that the behavior of some methods might change due to linguistic idiosyncrasies across different languages. We believe that this is a very important direction for future work. Replicating our experiments requires access to GPUs. 

\section*{Acknowledgements}

ZZ, KB and NA are supported by EPSRC grant EP/V055712/1, part of the European Commission CHIST-ERA programme, call 2019 XAI: Explainable Machine Learning-based Artificial Intelligence. This project made use of time on Tier 2 HPC facility JADE2, funded by EPSRC (EP/T022205/1).

\bibliography{anthology,custom}
\bibliographystyle{acl_natbib}

\clearpage

\appendix

\section{Data split statistics}\label{app:data_stat}

For X-FACT, FactCheck, AmazDigiMu and AmazPantry, we take the earliest 80\% data as the training and testing set for Syn, the earlier 80\% to 90\% data as Asy1 and the rest (i.e. the latest 10\%) as Asy2.
For Yelp and AGNews, which have more available data for each year, we sample a same number of data from each year based on the year with the least data available. This allows year-wise analysis. We take the newer 2 years of data as the Asy datasets, per year per Asy.
We also experiment with the original dataset (OSyn, including Syn, Asy1 and Asy2) to provide a comparison. Table \ref{tab:full_data_stat} shows details of each split for each task.

\begin{table*}[t!] 
\resizebox{\textwidth}{!}{%
\begin{tabular}{@{}cclrrcp{0.18\linewidth}p{0.18\linewidth}c@{}}
\toprule
{\bf Task} & {\bf Label Num} & {\bf Domain} & \multicolumn{1}{l}{{\bf Start Date}} & \multicolumn{1}{l}{{\bf End Date}} & {\bf Time Span(Days)} & \multicolumn{1}{l}{{\bf Median Date}} & {\bf Interquartile Time Span (Days)} & {\bf Data Num} \\ \midrule

\multirow{7}{*}{AGNews} & \multirow{7}{*}{4} 
    & OSyn Full & 2004-08-18 & 2008-02-20 & 1281 & 2006-09-27 & 980 & 46790 \\
 &  & OSyn Train & 2004-08-18 & 2008-02-20 & 1281 & 2006-09-27 & 979 & 28074 \\
 &  & OSyn Test & 2004-08-18 & 2008-02-20 & 1281 & 2006-09-28 & 983 & 9358 \\
 &  & Syn Train & 2004-08-18 & 2006-12-20 & 854 & 2005-01-20 & 648 & 9358 \\
 &  & Syn Test & 2004-08-18 & 2006-12-20 & 854 & 2005-01-20 & 650 & 9358 \\
 &  & Asy1 Test & 2007-01-30 & 2007-12-31 & 335 & 2007-06-29 & 191 & 9358 \\
 &  & Asy2 Test & 2008-01-01 & 2008-02-20 & 50 & 2008-01-24 & 25 & 9358 \\ \midrule
\multirow{7}{*}{X-FACT} & \multirow{7}{*}{6} 
    & OSyn Full & 2007-01-04 & 2018-11-12 & 4330 & 2013-06-27 & 1727 & 12050 \\
 &  & OSyn Train & 2007-01-04 & 2018-11-12 & 4330 & 2013-08-19 & 1769 & 9639 \\
 &  & OSyn Test & 2007-04-26 & 2018-11-07 & 4213 & 2013-09-19 & 1727 & 1206 \\
 &  & Syn Train & 1995-04-01 & 2016-08-31 & 7823 & 2012-08-30 & 1296 & 7232 \\
 &  & Syn Test & 2007-01-04 & 2016-08-31 & 3527 & 2012-08-17 & 1320 & 1204 \\
 &  & Asy1 Test & 2016-08-31 & 2017-09-30 & 395 & 2017-02-24 & 216 & 1205 \\
 &  & Asy2 Test & 2017-09-30 & 2018-11-12 & 408 & 2018-05-10 & 209 & 1204 \\ \midrule
 
\multirow{7}{*}{FactCheck} & \multirow{7}{*}{2} 
    & OSyn Full & 1995-09-25 & 2021-07-19 & 9429 & 2016-12-29 & 1519 & 12640 \\
 &  & OSyn Train & 1996-02-27 & 2021-07-19 & 9274 & 2016-12-29 & 1527 & 10086 \\
 &  & OSyn Test & 1995-09-25 & 2021-06-23 & 9403 & 2016-12-20 & 1515 & 1277 \\
 &  & Syn Train & 1995-09-25 & 2019-05-01 & 8619 & 2016-04-14 & 1437 & 7446 \\
 &  & Syn Test & 1996-08-02 & 2019-05-01 & 8307 & 2016-03-09 & 1319 & 1241 \\
 &  & Asy1 Test & 2019-05-02 & 2020-05-15 & 379 & 2019-11-18 & 182 & 1368 \\
 &  & Asy2 Test & 2020-05-15 & 2021-07-19 & 430 & 2020-10-12 & 132 & 1368 \\ \midrule
 
\multirow{7}{*}{AmazDigiMu} & \multirow{7}{*}{3} 
    & OSyn Full & 1998-07-09 & 2018-09-26 & 7384 & 2015-01-19 & 789 & 169623 \\
 &  & OSyn Train & 1998-07-09 & 2018-09-26 & 7384 & 2015-01-20 & 788 & 135698 \\
 &  & OSyn Test & 1998-08-21 & 2018-09-20 & 7335 & 2015-01-08 & 794 & 16962 \\
 &  & Syn Train & 1998-08-21 & 2016-05-07 & 6469 & 2014-09-20 & 673 & 101774 \\
 &  & Syn Test & 1998-12-20 & 2016-05-07 & 6351 & 2014-09-18 & 669 & 16963 \\
 &  & Asy1 Test & 2016-05-07 & 2016-12-30 & 237 & 2016-08-12 & 120 & 16962 \\
 &  & Asy2 Test & 2016-12-30 & 2018-09-26 & 635 & 2017-08-07 & 290 & 16962 \\ \midrule

\multirow{7}{*}{AmazPantry} & \multirow{7}{*}{3} 
    & OSyn Full & 2006-04-09 & 2018-10-04 & 4561 & 2016-09-27 & 485 & 137611 \\
 &  & OSyn Train & 2006-04-09 & 2018-09-28 & 4555 & 2016-09-27 & 486 & 110088 \\
 &  & OSyn Test & 2006-10-14 & 2018-10-04 & 4373 & 2016-09-24 & 474 & 13761 \\
 &  & Syn Train & 2006-04-28 & 2017-07-30 & 4111 & 2016-07-06 & 413 & 82566 \\
 &  & Syn Test & 2006-12-22 & 2017-07-30 & 3873 & 2016-07-08 & 406 & 13762 \\
 &  & Asy1 Test & 2017-07-30 & 2018-01-21 & 175 & 2017-10-16 & 92 & 13761 \\
 &  & Asy2 Test & 2018-01-21 & 2018-10-04 & 256 & 2018-04-12 & 94 & 13761 \\ \midrule
 
 \multirow{7}{*}{Yelp} & \multirow{7}{*}{5} 
    & OSyn Full & 2005-02-16 & 2022-01-19 & 6181 & 2014-01-02 & 3274 & 15372 \\
 &  & OSyn Train & 2005-02-16 & 2022-01-19 & 6181 & 2014-01-30 & 3277 & 11956 \\
 &  & OSyn Test & 2005-02-16 & 2022-01-19 & 6181 & 2013-09-18 & 3297 & 1708 \\
 &  & Syn Train & 2005-02-16 & 2018-12-31 & 5066 & 2012-01-16 & 2556 & 8540 \\
 &  & Syn Test & 2005-02-16 & 2018-12-24 & 5059 & 2011-10-22 & 2457 & 1708 \\
 &  & Asy1 Test & 2019-01-01 & 2020-12-31 & 730 & 2020-01-01 & 375 & 1708 \\
 &  & Asy2 Test & 2021-01-01 & 2022-01-19 & 383 & 2022-01-01 & 195 & 1708 \\ \bottomrule
\end{tabular}%
}
\caption{Summary of the original dataset and the chronological splits for each task.}
\label{tab:full_data_stat}
\end{table*}

\section{Models and Hyper-parameters}\label{app:hyperparameters}

\paragraph{Feature attributions}
We use BERT-base with pre-trained weights from the Huggingface library \citep{wolf-etal-2020-transformers}. We use the AdamW optimizer \citep{loshchilov2017decoupled} with an initial learning rate of $1e^{-5}$ for fine-tuning BERT and $1e^{-4}$ for the fully-connected classification layer. We train our models for 3 epochs using a linear scheduler, with 10\% of the data in the first epoch as warming up. We also use a grad-norm of 1.0. The model with the lowest loss on the development set is selected. All models are trained across 5 random seeds, and we report the average and standard deviation. 

\paragraph{FRESH}
For the rationale extractor, we use the same model for extracting rationales from feature attributions. For the classifier (trained only on the extracted rationales), we also use BERT-base with the same optimizer configuration and scheduler warm-up steps. We use a grad-norm of 1.0 and select the model with the lowest loss on the development set. We train across 5 random seeds for 5 epochs.

\paragraph{HardKUMA}
We use the 300-dimensional pre-trained GloVe embeddings from the 840B release \citep{pennington-etal-2014-glove} and keep them frozen. Similar to \citet{bastings-etal-2019-interpretable} and \citet{chrysostomou-aletras-2022-empirical}, we use a Bi-LSTM layer of 200-d for the rationale extractor.
We use the Adam optimizer \citep{loshchilov2017decoupled} with a learning rate between $1e^{-3}$ and $1e^{-5}$ and a weight decay of $1e^{-5}$. We also enforce a grad-norm of 5.0 and train for 20 epochs across 5 random seeds. Following \citet{guerreiro-martins-2021-spectra}, we select the model with the highest F1-macro score on the development set and tuning the Lagrangian relaxation algorithm parameters between $1e^{-2}$ and $1e^{-5}$.

\paragraph{SPECTRA}
Following \citet{guerreiro-martins-2021-spectra}, we take the 300-dimensional pre-trained GloVe embeddings from the 840B release \citep{pennington-etal-2014-glove} as word representations and keep them frozen. 
As \citet{guerreiro-martins-2021-spectra} suggested, results with Bi-LSTM layers were competitive with those with BERT reported in \citet{jain2020learning}. We, therefore, instantiate all encoder networks as Bi-LSTM layers of hidden size 200. We also use the AdamW optimizer \citep{loshchilov2017decoupled} for training SPECTRA. We use a learning rate $\in$ [$1e^{-3}$, $5e^{-4}$, $1e^{-4}$, $5e^{-5}$]  and $l_2$ regularization $\in$ [$1e^{-4}$, $1e^{-5}$] for the training. We also use a grad-norm of 5.0. We train all models for highlights extraction for a minimum of 3 epochs and maximum of 20 epochs. For matching extraction, we set the number of minimum epochs of 3 and maximum epochs of 10. We implement early stopping for the model training if F1 stop increasing over 5 epochs and for highlights extraction if F1 stop increasing over 3 epochs. Tributes are paid to \citet{guerreiro-martins-2021-spectra} for the published code published.

All experiments are run on a single NVIDIA Tesla V100 GPU.

\section{Post-hoc Explanations Faithfulness}
\label{app:posthoc_results}

\renewcommand{\arraystretch}{1.2}
\begin{table*}[!t]
    \centering
    \footnotesize 
    \setlength{\tabcolsep}{2.0pt}
    \resizebox{1.0\textwidth}{!}{%
    \begin{tabular}{p{0.09\textwidth}cp{0.09\textwidth}||p{0.06\textwidth}||cccccccc||cccccccc}
    \toprule
         \textbf{Task} & \textbf{Train} & \textbf{Test} & \textbf{Fulltext} &  \multicolumn{8}{c||}{\textbf{Normalised Sufficiency}} & \multicolumn{8}{c}{\textbf{Normalised Comprehensiveness}}\\
          & \textbf{Set} & \textbf{Set} & \textbf{F1} 
          & $\alpha\nabla\alpha$ & $\alpha$ & DL & $x\nabla x $ & lime & IG & DLsp & Gsp
          & $\alpha\nabla\alpha$ & $\alpha$ & DL & $x\nabla x $ & lime & IG & DLsp & Gsp \\\hline \hline
        & Original
        & Original & 90.8 & 1.53 & 1.34 & 1.00 & 1.31 & 0.99 & 1.32 & 1.04 & 1.00 & 2.42 & 2.09 & 1.23 & 1.87 & 1.20 & 1.88 & 0.98 & 0.96 \\ \cline{2-20}
        & & Syn & 89.1 & 1.51 & 1.36 & 0.93 & 1.32 & 0.98 & 1.31 & 0.98 & 0.97 & 2.95 & 2.18 & 1.36 & 2.00 & 1.22 & 1.97 & 1.01 & 1.03 \\ 
        AGNews & Syn & Asy1 & 87.5 & 1.48 & 1.37 & 0.97 & 1.29 & 1.06 & 1.29 & 1.02 & 0.97 & 2.96 & 2.35 & 1.36 & 2.00 & 1.43 & 1.97 & 1.03 & 1.01  \\ 
        & & Asy2 & 85.5 & 1.49 & 1.40 & 0.97 & 1.26 & 1.02 & 1.25 & 1.02 & 0.97 & 3.08 & 2.51 & 1.41 & 2.04 & 1.38 & 1.99 & 1.02 & 1.02 \\
        \hline
        & Original
        & Original & 37.4 & 1.24 & 1.12 & 0.88 & 1.14 & 1.04 & 1.15 & 0.95 & 0.98 & 1.67 & 1.40 & 1.12 & 1.35 & 1.06 & 1.37 & 1.00 & 0.98 \\ \cline{2-20}
        & & Syn & 38.2 & 1.16 & 1.02 & 0.91 & 1.02 & 1.03 & 1.03 & 0.93 & 0.95 & 1.64 & 1.02 & 1.04 & 1.22 & 1.05 & 1.19 & 0.90 & 0.96 \\
        X-FACT & Syn & Asy1 & 37.4 & 1.15 & 0.99 & 0.98 & 1.03 & 1.03 & 1.04 & 0.95 & 0.95 & 1.74 & 1.03 & 1.08 & 1.26 & 1.06 & 1.24 & 1.04 & 1.00 \\
        & & Asy2 & 38.2 & 1.14 & 0.98 & 0.96 & 1.02 & 1.03 & 1.02 & 0.94 & 0.95 & 1.58 & 0.92 & 1.00 & 1.15 & 1.01 & 1.10 & 0.93 & 0.92 \\
\hline 
        & Original
        & Original & 74.4 & 2.00 & 1.73 & 1.10 & 1.09 & 1.19 & 1.11 & 1.03 & 0.98 & 4.67 & 3.42 & 1.26 & 2.23 & 1.12 & 2.27 & 1.10 & 1.07\\ \cline{2-20}
        & & Syn & 71.1 & 2.8 & 2.4 & 1.17 & 1.41 & 1.06 & 1.48 & 1.13 & 1.03 & 2.85 & 2.67 & 1.52 & 1.95 & 0.96 & 1.96 & 1.16 & 1.08 \\
        FactCheck & Syn & Asy1 & 70.6 & 1.15 & 1.2 & 0.94 & 0.96 & 1.04 & 1.00 & 0.91 & 0.88 & 0.99 & 1.12 & 0.84 & 1.10 & 1.03 & 0.94 & 1.15 & 1.01 \\
        & & Asy2 & 73.5 & 1.26 & 1.13 & 1.00 & 1.22 & 1.24 & 1.23 & 1.08 & 0.96 & 1.06 & 1.18 & 0.83 & 1.00 & 1.00 & 0.91 & 1.04 & 0.98 \\
        \hline 
        & Original
        & Original & 71.4 & 3.19 & 2.97 & 1.81 & 1.89 & 0.58 & 1.96 & 1.45 & 1.07 & 1.61 & 1.45 & 1.16 & 1.21 & 0.63 & 1.19 & 1.03 & 1.13 \\ \cline{2-20}
        & & Syn & 69.3          & 1.31 & 1.25 & 0.51 & 0.83 & 0.55 & 0.71 & 0.95 & 0.77 & 2.71 & 1.93 & 1.55 & 1.76 & 1.08 & 2.11 & 0.92 & 1.19 \\
        AmazDigiMu & Syn & Asy1 & 63.7 & 1.19 & 1.15 & 0.48 & 0.84 & 0.57 & 0.74 & 0.93 & 0.81 & 2.63 & 1.74 & 1.23 & 1.82 & 1.09 & 2.28 & 0.80 & 1.19 \\
        & & Asy2 & 58.2 & 1.16 & 1.14 & 0.49 & 0.84 & 0.58 & 0.74 & 0.94 & 0.82 & 2.52 & 1.63 & 1.10 & 1.73 & 1.05 & 2.14 & 0.75 & 1.18 \\
        \hline 
        & Original
        & Original & 71.0 & 2.40 & 1.03 & 0.97 & 1.15 & 0.88 & 1.37 & 0.82 & 0.78 & 1.99 & 1.08 & 0.94 & 1.60 & 1.19 & 1.48 & 0.91 & 1.25\\ \cline{2-20}
        & & Syn & 69.6 & 2.18 & 1.55 & 0.82 & 1.47 & 0.57 & 1.71 & 0.82 & 0.97 & 2.14 & 1.68 & 1.03 & 1.62 & 0.76 & 1.55 & 0.77 & 0.85 \\
        AmazPantry & Syn & Asy1 & 68.2 & 2.14 & 1.54 & 0.81 & 1.51 & 0.59 & 1.73  & 0.79 & 1.00 & 2.13 & 1.63 & 1.00 & 1.62 & 0.77 & 1.55 & 0.75 & 0.87 \\
        & & Asy2 & 68.1 & 2.06 & 1.46 & 0.76 & 1.52 & 0.56 & 1.74 & 0.76 & 0.98 & 2.12 & 1.58 & 0.96 & 1.64 & 0.73 & 1.57 & 0.72 & 0.86 \\
\hline 
        & Original
        & Original & 61.4 & 2.58 & 1.73 & 1.16 & 1.62 & 0.97 & 1.60 & 1.03 & 0.99 & 2.27 & 1.56 & 1.13 & 1.64 & 0.76 & 1.61 & 1.05 & 1.01 \\\cline{2-20}
        & & Syn & 60.3 & 2.16 & 1.61 & 1.30 & 1.38 & 0.95 & 1.37 & 1.12 & 0.96 & 2.52 & 1.76 & 1.34 & 1.65 & 0.85 & 1.66 & 1.10 & 0.99 \\
        Yelp & Syn & Asy1 & 59.2 & 2.39 & 1.70 & 1.39 & 1.58 & 0.85 & 1.59 & 1.15 & 0.96 & 2.30 & 1.62 & 1.39 & 1.67 & 0.83 & 1.69 & 1.11 & 0.99 \\
        & & Asy2 & 61.0 & 2.39 & 1.69 & 1.36 & 1.60 & 0.88 & 1.61 & 1.13 & 0.95 & 2.27 & 1.60 & 1.31   & 1.63 & 0.80 & 1.65 & 1.07 & 0.95  \\
        \hline
    \end{tabular}}
    \caption{AOPC Normalized Sufficiency and Comprehensiveness (higher is better) for the OSyn, Syn and Asy of 8 feature attribution approaches on their TOPK tokens. Each feature is presented as the ratio to the random attribution baseline.} 
    \label{tab:predictive_and_faithfulness} 
\end{table*}

\section{Select-then-predict Predictive Performance}\label{app:sec:select_then_predict}

\subsection{HardKUMA and SPECTRA Predictive Performance}\label{app:sec:hardkuma_spectra}
\renewcommand{\arraystretch}{1.1}
\begin{table*}[h!]
    \setlength\tabcolsep{3.5pt}
    \centering
    \small
\begin{tabular}{ll||ccccccc}
\toprule
\multirow{2}{*}{Task} & \multirow{2}{*}{Domain} & \multicolumn{1}{l}{LSTM} & \multicolumn{1}{l}{LSTM} & \multicolumn{1}{l}{KUMA} & \multicolumn{1}{l}{KUMA} & \multicolumn{1}{l}{KUMA} & \multicolumn{1}{l}{SPECTRA} & \multicolumn{1}{l}{SPECTRA} \\ 

& & \multicolumn{1}{c}{F1} & \multicolumn{1}{c}{std} & \multicolumn{1}{c}{F1} & \multicolumn{1}{c}{std} & \multicolumn{1}{c}{len} & \multicolumn{1}{c}{F1} & \multicolumn{1}{c}{std} \\ \hline \hline

\multirow{4}{*}{AGNews} & OSyn& 86.8 & 0.1 & 85.0 & 0.4 & 23.2 & 83.2 & 0.4 \\ \cline{2-9}
 & Syn & 84.4 & 0.1 & 81.9 & 1.1 & 39.1 & 80.4 & 0.8 \\
 & Asy1 & 82.4 & 0.5 & 79.1 & 1.7 & 36.8 & 78.3 & 1.7 \\
 & Asy2 & 79.4 & 0.6 & 77.4 & 1.5 & 37.0 & 77.7 & 1.9 \\ \hline
 
\multirow{4}{*}{X-FACT} & OSyn& 35.4 & 1.5 & 11.7 & 1.8 & 41.1 & 24.9 & 1.1 \\ \cline{2-9}
 & Syn & 34.3 & 0.5 & 9.9 & 0.0 & 43.6 & 22.0 & 5.7 \\
 & Asy1 & 31.6 & 3.0 & 12.7 & 0.0 & 42.5 & 16.7 & 7.8 \\
 & Asy2 & 30.0 & 1.9 & 12.5 & 0.0 & 41.6 & 20.6 & 7.3 \\ \hline
 
\multirow{4}{*}{FactCheck} & OSyn& 62.4 & 2.9 & 47.2 & 1.7 & 53.9 & 50.7 & 5.3 \\ \cline{2-9}
 & Syn & 55.6 & 3.2 & 47.3 & 2.7 & 78.5 & 50.5 & 6.5 \\
 & Asy1 & 51.7 & 2.8 & 45.2 & 2.9 & 78.9 & 45.5 & 4.2 \\
 & Asy2 & 49.8 & 3.6 & 41.9 & 2.3 & 79.0 & 43.2 & 6.4 \\ \hline
 
\multirow{4}{*}{AmazDigiMu} & OSyn& 66.8 & 0.8 & 64.9 & 1.7 & 19.0 & 48.7 & 1.4 \\ \cline{2-9}
 & Syn & 66.4 & 1.0 & 65.8 & 1.7 & 18.2 & 43.0 & 2.9 \\
 & Asy1 & 59.1 & 0.9 & 56.9 & 1.3 & 18.7 & 37.6 & 3.6 \\
 & Asy2 & 54.4 & 1.2 & 52.1 & 1.2 & 18.6 & 39.0 & 3.6 \\ \hline
 
\multirow{4}{*}{AmazPantry} & OSyn& 67.5 & 0.4 & 64.4 & 0.4 & 18.4 & 58.1 & 0.5 \\ \cline{2-9}
 & Syn & 67.4 & 0.6 & 63.5 & 1.0 & 17.9 & 48.9 & 1.4 \\
 & Asy1 & 66.4 & 0.4 & 62.7 & 0.9 & 18.7 & 50.0 & 1.1 \\
 & Asy2 & 67.0 & 0.9 & 62.7 & 1.1 & 19.6 & 51.5 & 1.0 \\ \hline
 
\multirow{4}{*}{Yelp} & OSyn& 48.0 & 0.4 & 28.8 & 1.6 & 11.7 & 36.2 & 2.2 \\ \cline{2-9}
 & Syn & 39.1 & 1.3 & 19.4 & 0.5 & 11.3 & 32.8 & 6.5 \\
 & Asy1 & 43.3 & 1.4 & 20.1 & 0.4 & 14.9 & 33.6 & 4.4 \\
 & Asy2 & 43.2 & 1.8 & 20.0 & 0.8 & 15.0 & 34.7 & 5.8 \\ \bottomrule
\end{tabular}%
    \caption{Averaged macro F1 performance and standard deviation over five runs for HardKUMA and SPECTRA and their corresponding full-text models (T-test is conducted between HardKUMA and LSTM, and between SPECTRA and LSTM, for each split. 
    }
    \label{app:selective_rationalization}
\end{table*}

\subsection{FRESH Predictive Performance}\label{app:sec:all_fresh}
\renewcommand{\arraystretch}{1.1}
\begin{table*}[t!]
    \setlength\tabcolsep{3.5pt}
    \centering
    \small
\begin{tabular}{ll||cccccccc}
\toprule
\multirow{2}{*}{Task} & \multirow{2}{*}{Domain} & \multicolumn{1}{l}{BERT} & \multicolumn{1}{l}{BERT} & \multicolumn{1}{l}{$\alpha\nabla\alpha$} & \multicolumn{1}{l}{$\alpha\nabla\alpha$} & \multicolumn{1}{l}{DL} & \multicolumn{1}{l}{DL} & \multicolumn{1}{l}{$x\nabla x $} & \multicolumn{1}{l}{$x\nabla x $} \\ 

& & \multicolumn{1}{c}{F1} & \multicolumn{1}{c}{std} & \multicolumn{1}{c}{F1} & \multicolumn{1}{c}{std} & \multicolumn{1}{c}{F1} & \multicolumn{1}{c}{std} & \multicolumn{1}{c}{F1} & \multicolumn{1}{c}{std} \\ \hline \hline

\multirow{4}{*}{AGNews} & OSyn& 90.8 & 0.3 & 90.1 & 0.1 & 89.7 & 0.1 & 83.1 & 0.3 \\ \cline{2-10}
 & Syn & 89.1 & 0.5 & 87.0 & 0.6 & 86.7 & 0.4 & 81.5 & 0.6 \\
 & Asy1 & 87.5 & 0.7 & 84.6 & 0.8 & 83.6 & 0.9 & 75.0 & 1.3 \\
 & Asy2 & 85.5 & 0.5 & 83.3 & 0.3 & 81.6 & 0.9 & 73.7 & 1.7 \\ \hline
 
\multirow{4}{*}{X-FACT} & OSyn& 37.4 & 2.9 & 29.9 & 4.2 & 35.8 & 2.2 & 29.0 & 1.3 \\\cline{2-10}
 & Syn & 38.2 & 2.4 & 32.5 & 1.7 & 28.3 & 5.1 & 27.2 & 2.0 \\
 & Asy1 & 37.4 & 1.8 & 32.6 & 2.8 & 21.4 & 2.4 & 18.3 & 2.2 \\
 & Asy2 & 38.2 & 1.7 & 30.0 & 2.6 & 21.6 & 2.9 & 24.4 & 1.4 \\ \hline
 
\multirow{4}{*}{FactCheck} & OSyn& 74.4 & 3.2 & 77.1 & 0.4 & 83.9 & 0.3 & 64.2 & 3.0 \\\cline{2-10}
 & Syn & 71.1 & 1.9 & 71.8 & 0.2 & 64.7 & 1.0 & 53.2 & 4.3 \\
 & Asy1 & 70.6 & 2.7 & 72.1 & 0.0 & 67.0 & 1.3 & 45.9 & 2.4 \\
 & Asy2 & 73.5 & 1.7 & 74.2 & 0.1 & 65.7 & 1.5 & 42.8 & 2.7 \\\hline
 
\multirow{4}{*}{AmazDigiMu} & OSyn& 71.4 & 1.6 & 66.2 & 0.8 & 68.1 & 0.5 & 50.8 & 2.2 \\\cline{2-10}
 & Syn & 69.3 & 2.5 & 63.6 & 0.9 & 70.9 & 0.5 & 51.2 & 4.3 \\
 & Asy1 & 63.7 & 1.5 & 52.5 & 0.6 & 61.7 & 1.2 & 49.3 & 3.1 \\
 & Asy2 & 58.2 & 1.1 & 50.2 & 0.9 & 56.3 & 1.0 & 44.6 & 1.7 \\ \hline
 
\multirow{4}{*}{AmazPantry} & OSyn& 71.0 & 0.5 & 66.2 & 0.5 & 68.6 & 0.2 & 49.0 & 1.9 \\\cline{2-10}
 & Syn & 69.6 & 1.5 & 62.2 & 1.0 & 54.2 & 1.4 & 49.2 & 2.7 \\
 & Asy1 & 68.2 & 1.4 & 58.9 & 0.9 & 52.3 & 0.9 & 47.5 & 2.6 \\
 & Asy2 & 68.1 & 2.5 & 57.8 & 0.9 & 51.0 & 0.7 & 47.4 & 2.8 \\ \hline
 
\multirow{4}{*}{Yelp} & OSyn& 61.4 & 1.3 & 58.7 & 0.8 & 51.5 & 1.7 & 45.3 & 1.4 \\\cline{2-10}
 & Syn & 60.3 & 0.6 & 56.5 & 0.4 & 52.2 & 1.5 & 43.4 & 4.3 \\
 & Asy1 & 59.2 & 1.3 & 55.0 & 0.8 & 52.4 & 0.8 & 46.1 & 2.4 \\
 & Asy2 & 61.0 & 0.7 & 58.5 & 0.7 & 54.0 & 1.4 & 45.5 & 3.7 \\ \bottomrule
\end{tabular}%
    \caption{Averaged macro F1 performance and standard deviation over five runs for HardKUMA and SPECTRA and their corresponding full-text models.
    }
    \label{app:all_fresh}
\end{table*}

\end{document}